\documentclass[conference]{IEEEtran}
\IEEEoverridecommandlockouts

\usepackage{cite}
\usepackage{amsmath,amssymb,amsfonts}
\usepackage{graphicx}
\usepackage{textcomp}
\usepackage{xcolor}
\usepackage{booktabs}
\usepackage{multirow}
\usepackage{subcaption}
\usepackage{microtype}
\usepackage{url}
\usepackage{array}
\usepackage{algorithm}
\usepackage{algorithmic}
\usepackage{enumitem}

\usepackage{tikz}
\usepackage{pgfplots}
\pgfplotsset{compat=1.18}
\usepgfplotslibrary{groupplots}
\usetikzlibrary{
  arrows.meta,
  shapes.geometric,
  shapes.multipart,
  positioning,
  fit,
  calc,
  backgrounds,
  decorations.pathreplacing,
  decorations.markings,
  matrix
}

\usepackage[hidelinks]{hyperref}

\def\BibTeX{{\rm B\kern-.05em{\sc i\kern-.025em b}\kern-.08em
    T\kern-.1667em\lower.7ex\hbox{E}\kern-.125emX}}

\tikzset{
  block/.style={
    rectangle, rounded corners=3pt,
    draw=black!70, fill=#1,
    text width=1.9cm, align=center,
    minimum height=0.65cm, font=\footnotesize
  },
  block/.default=white,
  wideblock/.style={
    rectangle, rounded corners=3pt,
    draw=black!70, fill=#1,
    text width=2.5cm, align=center,
    minimum height=0.65cm, font=\footnotesize
  },
  wideblock/.default=white,
  tinyblock/.style={
    rectangle, rounded corners=2pt,
    draw=black!60, fill=#1,
    text width=1.5cm, align=center,
    minimum height=0.55cm, font=\scriptsize
  },
  tinyblock/.default=white,
  fpnblock/.style={
    rectangle, rounded corners=2pt,
    draw=black!70, fill=#1,
    minimum width=1.1cm, minimum height=0.45cm,
    align=center, font=\scriptsize
  },
  fpnblock/.default=white,
  arrow/.style={-{Stealth[length=5pt]}, thick, draw=black!70},
  darrow/.style={-{Stealth[length=4pt]}, draw=black!60},
  groupbox/.style={
    rectangle, rounded corners=5pt,
    draw=#1!60, fill=#1!8,
    inner sep=6pt
  },
  groupbox/.default=blue
}

\begin{document}

\title{TEEP-RCNN: Texture-Enhanced Edge-aware Perception\\
  for Steel Surface Defect Detection via\\
  Improved Convolutional Block Attention in Faster R-CNN}

\author{
  \IEEEauthorblockN{Kirtan Rajesh}
  \IEEEauthorblockA{
    \textit{AI Researcher and Engineer}\\
    \textit{Independent Researcher}\\
    kirtanrajesh@gmail.com
  }
}

\maketitle

\begin{abstract}
Steel surface defect detection is critical for automated industrial quality control but remains challenging due to subtle inter-class texture differences and pronounced class imbalance. We introduce TEEP-RCNN (Texture-Enhanced Edge-aware Perception Region-based CNN), a two-stage detector built on Faster R-CNN with a Feature Pyramid Network backbone and an improved Convolutional Block Attention Module (CBAM). Our CBAM adds dropout regularization in the channel attention MLP and batch normalization on the spatial attention branch, reducing co-adaptation and stabilizing gating logits. Training uses a differential learning rate protocol with cosine annealing warm-up, separating update rates for the pre-trained ResNet-101 backbone and the detection head. At inference, predictions are refined via Test-Time Augmentation fused with Weighted Box Fusion (WBF), improving localization stability on elongated and boundary-adjacent defects. On the NEU-DET benchmark across six defect categories, TEEP-RCNN achieves 73.3\% mAP@50 and 37.9\% mAP@50-95 in only 10 training epochs on a single GPU, competitive with YOLOv11m (76.2\% mAP@50, 100 epochs) while outperforming it on the rolled-in-scale category under the COCO metric. Per-class analysis shows the spatial attention branch is most effective on elongated texture defects such as patches and scratches, while crazing remains an open challenge across both paradigms due to its distributed non-local texture structure.
\end{abstract}


\section{Introduction}

Hot-rolled steel strip is one of the most widely manufactured structural materials globally, with annual production volumes exceeding one billion metric tons. Surface defects introduced during the rolling, cooling, and descaling stages, such as crazing, inclusions, patches, pitted surfaces, rolled-in scale, and scratches, represent a persistent quality problem. Industry estimates suggest that surface defects account for a significant fraction of downstream manufacturing rejects and rework costs, and that even a modest improvement in detection accuracy at the strip inspection stage can prevent material wastage and reduce recalls of formed components. Traditional online inspection relies on human operators visually monitoring steel strips at line speeds between 5 and 30 meters per second, a task where fatigue-induced missed detections are well documented. Automated machine vision systems therefore represent a practical necessity rather than an optional optimization.

Early automated methods used hand-crafted feature extractors tailored to specific defect morphologies. Local Binary Patterns (LBP) \cite{song2013noise}, Gabor filter banks, and wavelet-based texture descriptors capture periodic and directional surface structure and have been shown to discriminate several defect types under controlled conditions. However, these representations require separate parameter tuning for each defect category and mill configuration, and they fail systematically when the illumination conditions or surface finish distribution shifts from the calibration environment. This fragility motivates learning-based approaches that derive representations directly from annotated examples.

Deep convolutional neural networks introduced a paradigm shift in this space. Two-stage detectors based on Faster R-CNN \cite{ren2015faster}, which pairs a Region Proposal Network (RPN) with a classification-regression head over a shared convolutional backbone, achieved strong localization accuracy on natural image benchmarks and transferred well to industrial inspection tasks via fine-tuning. Feature Pyramid Networks (FPN) \cite{lin2017feature} strengthened this pipeline by constructing a top-down multi-scale feature hierarchy, allowing the detection head to simultaneously leverage high spatial resolution from shallow layers and strong semantic content from deep layers. Single-stage detectors such as the YOLO family \cite{jocher2023yolov8} removed the proposal generation stage and demonstrated competitive accuracy at substantially higher throughput, making them attractive for real-time deployment.

Despite these advances, a key gap remains: standard two-stage detectors apply the same generic feature representations to all object categories without differentiating between the distinct texture signatures of different defect types. Crazing, for example, produces a fine-grained crack network that activates across many spatial locations, while scratches produce a narrow high-contrast linear response that is spatially concentrated. A detector that treats these patterns identically at the feature level will necessarily spend proposal capacity and classification confidence on irrelevant background regions.

Attention mechanisms offer a principled solution. The Convolutional Block Attention Module (CBAM) \cite{woo2018cbam} applies learned channel and spatial recalibration within each convolutional block, allowing the network to suppress uninformative feature channels and focus spatial activation on defect-relevant regions. However, the original CBAM design does not include regularization within the channel attention MLP, which can lead to over-reliance on a small subset of dominant channels, and it applies the spatial attention convolution without normalization, which can produce saturated binary gates early in training. Both issues are particularly consequential on a small-scale industrial dataset like NEU-DET, where the training set contains only 1,224 images after validation splitting.

This paper presents \textbf{TEEP-RCNN}, which addresses both gaps through two minimal but measurable modifications to the CBAM module, combined with a carefully designed training and inference strategy. The improved CBAM inserts a Dropout layer with rate 0.1 between the compression and expansion layers of the channel attention MLP, and adds Batch Normalization after the spatial attention convolution. These changes add a negligible number of parameters while providing the regularization needed to train a robust attention module on limited industrial data. The training protocol uses differential learning rates with cosine annealing warm-up, and the inference pipeline fuses predictions across test-time augmented views using Weighted Box Fusion.

\noindent The primary contributions of this paper are:
\begin{enumerate}[leftmargin=*]
  \item An \textbf{improved CBAM} with dropout in the channel attention MLP and batch normalization in the spatial attention branch, applied to the Layer4 bottleneck blocks of a ResNet-101 backbone within a Faster R-CNN two-stage detector.
  \item A \textbf{differential learning rate} training protocol combining AdamW with distinct learning rates for the pre-trained backbone and the task-specific detection head, preventing feature degradation in ImageNet-pretrained weights.
  \item A \textbf{WBF-TTA inference pipeline} that fuses predictions over test-time augmented views using Weighted Box Fusion, improving localization consistency on elongated defects with variable aspect ratios.
  \item A comprehensive \textbf{comparative evaluation} on NEU-DET against YOLOv8l, YOLOv11m, and a vanilla Faster R-CNN baseline, including epoch-by-epoch convergence profiling, per-class breakdown, and model efficiency analysis.
\end{enumerate}

The remainder of the paper is organized as follows. Section~\ref{sec:related} reviews related work. Section~\ref{sec:dataset} describes the dataset and augmentation strategy. Section~\ref{sec:method} details the proposed method. Section~\ref{sec:experiments} presents experimental results. Section~\ref{sec:discussion} analyzes findings. Section~\ref{sec:limitations} states limitations. Section~\ref{sec:conclusion} concludes.

\section{Related Work}
\label{sec:related}

\subsection{Traditional Defect Detection}
Early vision-based inspection systems built representations from image processing primitives. Morphological operations combined with adaptive thresholding identified macro-scale defects on relatively uniform backgrounds but were not robust to surface reflectance variation. Song and Yan \cite{song2013noise} proposed a completed LBP descriptor for hot-rolled steel and demonstrated improved illumination robustness by encoding both sign and magnitude transitions of the local texture pattern. Gabor filter banks at multiple orientations captured directional texture energy and found application in distinguishing directional defects such as scratches from isotropic patterns such as pitting. Despite the elegance of these interpretable descriptors, they require separate calibration per defect type and per production line, and their fixed feature vocabulary offers no mechanism for adaptation when the data distribution shifts.

\subsection{Two-Stage CNN Detectors}
Girshick et al.\ introduced R-CNN \cite{girshick2014rcnn}, the first method to apply CNN features to object proposals derived from selective search, achieving a substantial accuracy improvement over prior art on the PASCAL VOC benchmark. Fast R-CNN \cite{girshick2015fastrcnn} eliminated redundant convolutions by computing features once over the full image and projecting proposal regions via RoI pooling. Faster R-CNN \cite{ren2015faster} replaced selective search with a convolutional RPN that shares features with the detection head, enabling end-to-end training and reducing inference time to near real-time. Feature Pyramid Networks \cite{lin2017feature} added lateral connections and a top-down pathway to the backbone, constructing a multi-resolution feature hierarchy that captures both fine-grained edge information in shallow layers and semantically rich representations in deep layers. Mask R-CNN \cite{he2017maskrcnn} extended this family with a pixel-level segmentation branch and introduced RoI Align, replacing the quantized RoI pooling with bilinear interpolation for improved spatial precision. These developments collectively established the Faster R-CNN family as the dominant two-stage detection framework and motivated its application to industrial inspection.

\subsection{Single-Stage Detectors}
SSD \cite{liu2016ssd} removed the proposal stage by predicting class probabilities and bounding box offsets directly from multiple feature map scales, achieving a favorable accuracy-speed trade-off. RetinaNet \cite{lin2017focal} introduced focal loss to address the severe foreground-background class imbalance inherent in dense single-stage detection, down-weighting easy negatives so that rare, hard foreground instances dominate the gradient. The YOLO line \cite{jocher2023yolov8} evolved through numerous architectural revisions. YOLOv8 introduced an anchor-free detection head with decoupled classification and regression branches, a reparameterized convolutional stem, and a refined Mosaic augmentation strategy. YOLOv11 further improved the backbone with C3k2 blocks and SPPF pooling, and the medium variant (YOLOv11m) achieves 76.2\% mAP@50 on NEU-DET with only 20 million parameters and 67.7 GFLOPs at inference.

\subsection{Attention Mechanisms in Detection}
Hu et al.\ proposed SENet \cite{hu2018squeeze}, which recalibrates channel-wise responses by learning a global descriptor via global average pooling and a squeeze-and-excitation MLP. This module demonstrated consistent improvements when inserted into residual backbone architectures. BAM \cite{park2018bam} combined channel and spatial attention in a parallel bottleneck design and added the result as a residual to the feature map, providing both local and global refinement. CBAM \cite{woo2018cbam} applied channel and spatial attention sequentially: first recalibrating feature channels based on global statistics, then refining the spatial activation map using a 7$\times$7 convolutional gate. The sequential design is computationally lightweight and showed consistent gains over the base architecture on both classification and detection benchmarks including MS COCO. Non-local networks \cite{wang2018nonlocal} extended attention to long-range spatial dependencies through self-attention computation, at higher cost.

Our work directly extends CBAM by identifying two regularization weaknesses, specifically the absence of dropout in the channel MLP and the absence of normalization in the spatial branch, and correcting them with minimal architectural change. This improves training stability on small datasets without altering the fundamental design philosophy of sequential lightweight attention.

\subsection{Transformer-Based and Anchor-Free Detectors}
DETR \cite{carion2020detr} replaced the detection head and post-processing pipeline with a Transformer encoder-decoder and set-based bipartite matching, eliminating anchors and NMS entirely. Subsequent variants such as Deformable DETR \cite{zhu2021deformable} improved convergence speed by restricting cross-attention to a small set of sampled reference points, making transformer-based detection practical at standard training budgets. Swin Transformer \cite{liu2021swin} introduced a hierarchical vision backbone with shifted-window self-attention that computes local attention at multiple scales, enabling its use as a drop-in replacement for CNN backbones in two-stage pipelines. On NEU-DET, transformer-based approaches offer strong performance at the cost of larger training budgets and typically require larger datasets for their attention layers to converge meaningfully. DSAT \cite{dsat2025} reports 83.1\% mAP@50 on NEU-DET using a dynamic sparse attention mechanism combined with a transformer decoder, establishing the current benchmark state of the art. Our approach achieves 73.3\% mAP@50 with a two-stage CNN-based model trained for 10 epochs, which represents a substantially more accessible computational budget.

\subsection{Deep Learning for Surface Defect Detection}
Tao et al.\ \cite{tao2018metallic} demonstrated that CNN-based architectures can reliably classify and localize metallic surface defects across diverse categories when trained with appropriate augmentation, establishing an early baseline for the CNN era of industrial inspection. Bergmann et al.\ \cite{bergmann2019mvtec} introduced MVTec AD, a widely adopted anomaly detection benchmark covering 15 industrial object and texture categories, driving advances in unsupervised defect detection and segmentation. Tabernik et al.\ \cite{tabernik2020segmentation} proposed a segmentation-based framework that trains end-to-end on weak annotations and demonstrated precise defect boundary localization on the DAGM dataset. He et al.\ \cite{he2020neudet} introduced NEU-DET and showed that hierarchical feature fusion within Faster R-CNN outperforms single-scale detectors for the six-class steel defect task. Subsequent work on NEU-DET has explored deformable convolutions \cite{dai2017deformable} in the RoI head to improve geometric alignment, graph convolutional reasoning over detected regions, and data augmentation strategies specifically designed for periodic texture patterns. A recurring finding across these works is that crazing and rolled-in scale are consistently harder to detect than patches and scratches, a result our experiments confirm and attribute to the non-local distributed nature of the former defect types.

\section{Dataset and Preprocessing}
\label{sec:dataset}

\subsection{NEU-DET Dataset}
The NEU-DET surface defect database \cite{he2020neudet} provides images of hot-rolled steel strips annotated with axis-aligned bounding boxes across six defect categories: crazing (Cr), inclusion (In), patches (Pa), pitted surface (PS), rolled-in scale (RS), and scratches (Sc). Images are captured under industrial line-scan conditions at 200$\times$200 pixels with controlled but variable illumination. We resize all images to 640$\times$640 pixels to align with the anchor scales and the ImageNet pre-training resolution of the backbone. The dataset is split into 1,440 training images and 360 test images; we further hold out 15\% of training images as a validation set, yielding 1,224 training, 216 validation, and 360 test images.

Table~\ref{tab:dataset} shows the per-class bounding box counts in the training split. The dataset exhibits notable imbalance: the inclusion class provides 856 instances while pitted surface has only 351, a ratio of approximately 2.4:1. For context, the six defect categories span distinct morphological regimes. Crazing produces a network of fine surface cracks distributed uniformly across the image. Inclusions are compact, rounded dark spots embedded in the surface. Patches are large irregular regions of surface roughness with diffuse boundaries. Pitted surfaces show small, dense circular cavities arranged in irregular clusters. Rolled-in scale manifests as elongated irregular flakes partially pressed into the surface. Scratches are narrow, high-contrast linear tracks of varying length and orientation.

The bounding box statistics reflect these morphological differences. The mean bounding box area is 6,868 pixels$^2$ with a median of 4,619 pixels$^2$ on the 640$\times$640 scale. The mean aspect ratio across all classes is 1.06 with a median of 0.64, indicating a slight skew toward horizontally elongated boxes, driven by scratch and rolled-in-scale instances. Under the COCO object size taxonomy, all boxes in this dataset fall in the medium-to-large range (area $>$ 32$^2$ pixels), which explains why the size-stratified metrics in Section~\ref{sec:experiments} report non-zero values only for medium and large categories.

\begin{table}[t]
  \centering
  \caption{NEU-DET Instance Distribution (Training Split)}
  \label{tab:dataset}
  \renewcommand{\arraystretch}{1.15}
  \begin{tabular}{lccc}
    \toprule
    \textbf{Class} & \textbf{Instances} & \textbf{Fraction (\%)} & \textbf{Morphology} \\
    \midrule
    Crazing          & 537 & 15.7 & Distributed cracks \\
    Inclusion        & 856 & 25.1 & Compact spots \\
    Patches          & 674 & 19.8 & Diffuse regions \\
    Pitted Surface   & 351 & 10.3 & Dense cavities \\
    Rolled-in Scale  & 506 & 14.8 & Elongated flakes \\
    Scratches        & 420 & 12.3 & Linear tracks \\
    \midrule
    \textbf{Total}   & \textbf{3,410} & \textbf{100.0} & -- \\
    \bottomrule
  \end{tabular}
\end{table}

\subsection{Data Augmentation}
We use Albumentations \cite{buslaev2020albumentations} for all augmentation operations, applied only during training with bounding box coordinates transformed consistently. Table~\ref{tab:augmentation} details the full pipeline. Geometric augmentations include flips, 90-degree rotations, and shift-scale-rotate transforms, which address the rotational and positional symmetry of defects during manufacturing. Photometric augmentations, including brightness and contrast jitter, gamma randomization, and CLAHE, handle illumination variability common in line-scan cameras. CLAHE with a clip limit of 3.0 is particularly important for pitted surface instances, where low-contrast cavity boundaries benefit from adaptive local contrast enhancement.

Structural augmentations such as elastic transform and grid distortion simulate the surface deformation that can occur under different rolling tensions. Coarse dropout and grid dropout train the model to detect partially occluded defects, which arise when one defect overlaps another in dense surface regions. A minimum bounding box visibility threshold of 0.3 ensures that boxes with less than 30\% of their area remaining after spatial transforms are discarded rather than retained as noise.

All images are normalized with ImageNet mean and standard deviation, consistent with the ResNet-101 pre-training regime. For the extended 200-epoch run, we additionally apply class-conditional loss weights: crazing receives weight 3.0, rolled-in scale 2.5, and scratches 1.5, with the remaining classes at the default weight of 1.0, to partially offset the instance imbalance described above.

\begin{table}[t]
  \centering
  \caption{Training Augmentation Pipeline}
  \label{tab:augmentation}
  \renewcommand{\arraystretch}{1.1}
  \begin{tabular}{lcc}
    \toprule
    \textbf{Transform} & \textbf{Probability} & \textbf{Key Parameters} \\
    \midrule
    Horizontal Flip         & 0.50 & -- \\
    Vertical Flip           & 0.50 & -- \\
    Random Rotate 90        & 0.50 & -- \\
    Shift-Scale-Rotate      & 0.70 & shift=0.1, scale=0.15, rot=15\textdegree \\
    Brightness/Contrast     & 0.60 & $\pm$0.2 \\
    Random Gamma            & 0.40 & (80, 120) \\
    CLAHE                   & 0.50 & clip=3.0, tile=(8,8) \\
    Gaussian/Motion Blur    & 0.30 & limit=3 \\
    Gaussian Noise          & 0.30 & var=(5, 25) \\
    Random Shadow           & 0.30 & -- \\
    Coarse Dropout          & 0.40 & 8 holes, 32$\times$32 \\
    Grid Distortion         & 0.20 & steps=5, limit=0.1 \\
    Elastic Transform       & 0.20 & $\alpha$=1, $\sigma$=50 \\
    Grid Dropout            & 0.20 & -- \\
    \midrule
    Normalize               & 1.00 & ImageNet mean/std \\
    \bottomrule
  \end{tabular}
\end{table}

\section{Proposed Method: TEEP-RCNN}
\label{sec:method}

\subsection{Architecture Overview}

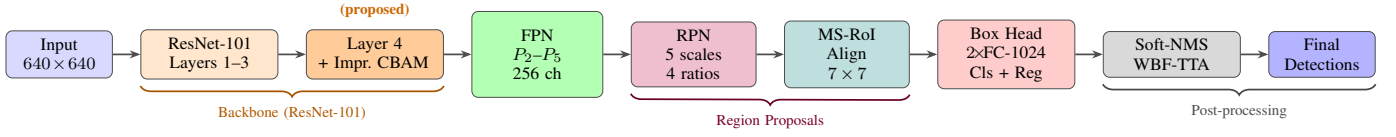
\begin{figure*}[t]
  \centering
  \resizebox{\linewidth}{!}{%
  \begin{tikzpicture}[node distance=0cm and 0.45cm]

    \node[block=blue!15, text width=1.5cm] (input) {Input\\$640\!\times\!640$};
    \node[block=orange!20, text width=2.0cm, right=0.45cm of input] (layer1)
      {ResNet-101\\Layers 1--3};
    \node[block=orange!40, text width=2.0cm, right=0.45cm of layer1] (layer4)
      {Layer 4\\+ Impr.\ CBAM};
    \node[block=green!30, text width=1.9cm, minimum height=1.4cm, right=0.45cm of layer4] (fpn)
      {FPN\\$P_2$--$P_5$\\256 ch};
    \node[block=purple!25, text width=1.8cm, right=0.45cm of fpn] (rpn)
      {RPN\\5 scales\\4 ratios};
    \node[block=teal!25, text width=1.8cm, right=0.45cm of rpn] (roi)
      {MS-RoI\\Align\\$7\!\times\!7$};
    \node[block=red!20, text width=2.0cm, right=0.45cm of roi] (head)
      {Box Head\\$2\!\times\!$FC-1024\\Cls + Reg};
    \node[block=gray!30, text width=2.0cm, right=0.45cm of head] (nms)
      {Soft-NMS\\WBF-TTA};
    \node[block=blue!30, text width=1.5cm, right=0.45cm of nms] (out)
      {Final\\Detections};

    \foreach \from/\to in {input/layer1, layer1/layer4, layer4/fpn,
                           fpn/rpn, rpn/roi, roi/head, head/nms, nms/out}
      \draw[arrow] (\from) -- (\to);

    \draw[decorate, decoration={brace, amplitude=5pt, mirror, raise=3pt},
          thick, orange!65!black]
      (layer1.south west) -- (layer4.south east)
      node[midway, below=9pt, font=\scriptsize, text=orange!70!black]
        {Backbone (ResNet-101)};

    \draw[decorate, decoration={brace, amplitude=5pt, mirror, raise=3pt},
          thick, purple!55!black]
      (rpn.south west) -- (roi.south east)
      node[midway, below=9pt, font=\scriptsize, text=purple!65!black]
        {Region Proposals};

    \draw[decorate, decoration={brace, amplitude=5pt, mirror, raise=3pt},
          thick, gray!55!black]
      (nms.south west) -- (out.south east)
      node[midway, below=9pt, font=\scriptsize, text=gray!60!black]
        {Post-processing};

    \node[font=\scriptsize, text=orange!80!black, above=0.07cm of layer4]
      {\textbf{(proposed)}};

  \end{tikzpicture}}%
  \vspace{2pt}
  \caption{TEEP-RCNN forward pipeline. The improved CBAM is injected into the Layer~4 bottleneck blocks of ResNet-101. FPN constructs a four-level feature pyramid fed to both the RPN and the RoI head. Soft-NMS and WBF-TTA post-process the final predictions at inference time.}
  \label{fig:architecture}
\end{figure*}

TEEP-RCNN is built on the Faster R-CNN two-stage detection framework \cite{ren2015faster} with three targeted additions: an improved CBAM module inserted into the ResNet-101 backbone, an anchor configuration tuned to the morphology of steel defects, and a WBF-TTA inference pipeline. Figure~\ref{fig:architecture} shows the complete forward pass.

Given an input image $\mathbf{I} \in \mathbb{R}^{3 \times H \times W}$ with $H = W = 640$, the backbone produces a hierarchy of feature maps at four spatial resolutions. The FPN aggregates these into four pyramid levels $\{P_2, P_3, P_4, P_5\}$, each with 256 channels, covering strides of $\{4, 8, 16, 32\}$ relative to the input. The RPN generates class-agnostic proposals from each level by applying a shared $3 \times 3$ convolutional head. Proposals are mapped to $7 \times 7$ feature grids via Multi-Scale RoI Align \cite{he2017maskrcnn}, and the box head produces classification scores and regression offsets. At inference, Soft-NMS and WBF over TTA views refine the final detections.

\subsection{ResNet-101 Backbone with FPN}
\label{subsec:backbone}

The backbone is a ResNet-101 \cite{he2016deep} pre-trained on ImageNet \cite{deng2009imagenet}, with all five residual stage groups set trainable. ResNet-101 provides 23 bottleneck blocks in Layer4 (versus 6 in ResNet-50), giving the CBAM module more activation sites to modulate and a richer semantic feature space at the deepest level.

\begin{figure}[htbp]
  \centering
  \resizebox{0.92\columnwidth}{!}{%
  \begin{tikzpicture}[node distance=0.36cm and 0.5cm]

    \node[fpnblock=orange!15, minimum width=1.2cm] (c2) {$C_2$\;H/4};
    \node[fpnblock=orange!25, minimum width=1.2cm, above=0.36cm of c2] (c3) {$C_3$\;H/8};
    \node[fpnblock=orange!38, minimum width=1.2cm, above=0.36cm of c3] (c4) {$C_4$\;H/16};
    \node[fpnblock=orange!55, minimum width=1.2cm, minimum height=0.55cm,
          above=0.36cm of c4] (c5) {$C_5$\;H/32\\+CBAM};

    \node[fpnblock=green!28, minimum width=1.0cm, right=0.5cm of c5] (lat5) {$1\!\times\!1$};
    \node[fpnblock=green!28, minimum width=1.0cm, right=0.5cm of c4] (lat4) {$1\!\times\!1$};
    \node[fpnblock=green!28, minimum width=1.0cm, right=0.5cm of c3] (lat3) {$1\!\times\!1$};
    \node[fpnblock=green!28, minimum width=1.0cm, right=0.5cm of c2] (lat2) {$1\!\times\!1$};

    \node[fpnblock=blue!32, minimum width=1.0cm, right=0.5cm of lat5] (p5) {$P_5$\\256};
    \node[fpnblock=blue!24, minimum width=1.0cm, right=0.5cm of lat4] (p4) {$P_4$\\256};
    \node[fpnblock=blue!17, minimum width=1.0cm, right=0.5cm of lat3] (p3) {$P_3$\\256};
    \node[fpnblock=blue!10, minimum width=1.0cm, right=0.5cm of lat2] (p2) {$P_2$\\256};

    \node[fpnblock=teal!35, minimum width=0.9cm, right=0.42cm of p5] (o5) {3$\!\times\!$3};
    \node[fpnblock=teal!28, minimum width=0.9cm, right=0.42cm of p4] (o4) {3$\!\times\!$3};
    \node[fpnblock=teal!21, minimum width=0.9cm, right=0.42cm of p3] (o3) {3$\!\times\!$3};
    \node[fpnblock=teal!14, minimum width=0.9cm, right=0.42cm of p2] (o2) {3$\!\times\!$3};

    \foreach \a/\b in {c2/c3, c3/c4, c4/c5}
      \draw[darrow] (\a) -- (\b);

    \foreach \c/\l in {c5/lat5, c4/lat4, c3/lat3, c2/lat2}
      \draw[darrow] (\c) -- (\l);

    \draw[darrow] (lat5) -- (p5);
    \draw[darrow] (lat4.east) -- (p4.west);
    \draw[darrow] (lat3.east) -- (p3.west);
    \draw[darrow] (lat2.east) -- (p2.west);

    \draw[darrow, dashed] (p5.south)
      -- node[right, font=\tiny]{${\uparrow}{\times}2$} (p4.north);
    \draw[darrow, dashed] (p4.south)
      -- node[right, font=\tiny]{${\uparrow}{\times}2$} (p3.north);
    \draw[darrow, dashed] (p3.south)
      -- node[right, font=\tiny]{${\uparrow}{\times}2$} (p2.north);

    \foreach \p/\o in {p5/o5, p4/o4, p3/o3, p2/o2}
      \draw[darrow] (\p) -- (\o);

    \node[font=\tiny, left=0.1cm of c2, text=gray!60] {L1};
    \node[font=\tiny, left=0.1cm of c3, text=gray!60] {L2};
    \node[font=\tiny, left=0.1cm of c4, text=gray!60] {L3};
    \node[font=\tiny, left=0.1cm of c5, text=orange!70!black] {\textbf{L4}};

    \node[font=\tiny, text=gray!55, below=0.22cm of c2] {Backbone};
    \node[font=\tiny, text=gray!55, below=0.22cm of lat2] {Lateral};
    \node[font=\tiny, text=blue!55, below=0.22cm of p2] {FPN levels};
    \node[font=\tiny, text=teal!60, below=0.22cm of o2] {Output};

  \end{tikzpicture}}%
  \caption{Feature Pyramid Network construction in TEEP-RCNN. Bottom-up path (solid) builds a feature hierarchy; the improved CBAM is applied at Layer~4 ($C_5$). Lateral $1\times1$ projections normalize each level to 256 channels. The top-down dashed path adds $\times2$ upsampled features to shallower lateral maps. Each $P_2$--$P_5$ passes a $3\times3$ conv before feeding the RPN and RoI head.}
  \label{fig:fpn}
\end{figure}
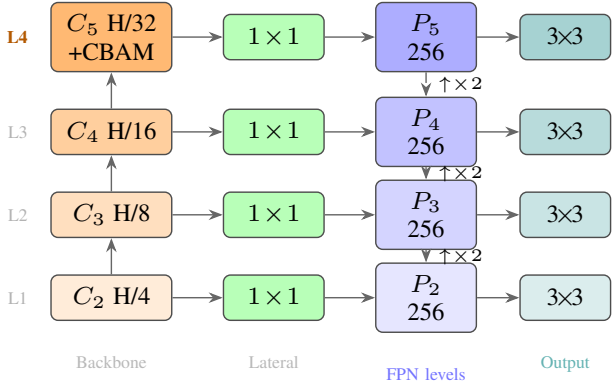

The FPN \cite{lin2017feature} builds a top-down pathway by upsampling the Layer4 output and adding it to Layer3 via a lateral $1 \times 1$ convolution, then repeating this for Layer2 and Layer1. This produces pyramid levels $P_2$ through $P_5$, each with 256 channels. $P_2$ carries the finest spatial resolution (stride 4) and is most useful for detecting small defect details, while $P_5$ carries the coarsest but most semantically meaningful representation (stride 32) useful for coarse defect localization and size estimation. The FPN design is visualized in Figure~\ref{fig:fpn}.

\subsection{Improved CBAM Module}
\label{subsec:cbam}

\begin{figure}[htbp]
  \centering
  \resizebox{0.92\columnwidth}{!}{%
  \begin{tikzpicture}[node distance=0.24cm and 0.35cm]

    \node[wideblock=gray!10] (inp) {Input $\mathbf{F}$\;$(C\!\times\!H\!\times\!W)$};

    \node[tinyblock=orange!20, below left=0.42cm and 0.75cm of inp] (avgpool) {AvgPool\\$\to 1\!\times\!1$};
    \node[tinyblock=orange!20, below right=0.42cm and 0.75cm of inp] (maxpool) {MaxPool\\$\to 1\!\times\!1$};

    \node[tinyblock=orange!30, below=0.24cm of avgpool] (mlp_a) {Shared\\MLP($\div$16)};
    \node[tinyblock=orange!30, below=0.24cm of maxpool] (mlp_m) {Shared\\MLP($\div$16)};

    \node[tinyblock=orange!45, below=0.24cm of mlp_a] (drop_a) {Dropout\\(0.1)};
    \node[tinyblock=orange!45, below=0.24cm of mlp_m] (drop_m) {Dropout\\(0.1)};

    \node[wideblock=orange!55] (ch_sig) at ($(drop_a)!0.5!(drop_m) + (0,-0.92)$)
      {$\sigma$(sum)$\to\mathbf{M}_c$\\channel gate};

    \node[wideblock=gray!20, below=0.42cm of ch_sig] (fprime)
      {$\mathbf{F}' = \mathbf{M}_c \otimes \mathbf{F}$};

    \node[tinyblock=teal!20, below left=0.42cm and 0.5cm of fprime] (sp_avg) {Ch-Avg\\pool};
    \node[tinyblock=teal!20, below right=0.42cm and 0.5cm of fprime] (sp_max) {Ch-Max\\pool};

    \node[wideblock=teal!25] (concat) at ($(sp_avg)!0.5!(sp_max) + (0,-0.88)$)
      {Concat $\to 2\!\times\!H\!\times\!W$};
    \node[wideblock=teal!35, below=0.24cm of concat] (conv77) {Conv2d $7\!\times\!7$};
    \node[wideblock=teal!50, below=0.24cm of conv77] (bn) {BatchNorm \textit{(new)}};
    \node[wideblock=teal!60, below=0.24cm of bn] (sp_sig) {$\sigma\to\mathbf{M}_s$\\spatial gate};

    \node[wideblock=blue!20, below=0.42cm of sp_sig] (out)
      {$\mathbf{F}'' = \mathbf{M}_s \otimes \mathbf{F}'$};

    \draw[darrow] (inp.south) -- ++(0,-0.18) -| (avgpool.north);
    \draw[darrow] (inp.south) -- ++(0,-0.18) -| (maxpool.north);
    \draw[darrow] (avgpool) -- (mlp_a);
    \draw[darrow] (maxpool) -- (mlp_m);
    \draw[darrow] (mlp_a) -- (drop_a);
    \draw[darrow] (mlp_m) -- (drop_m);
    \draw[darrow] (drop_a.south) to[out=270, in=135] (ch_sig.north);
    \draw[darrow] (drop_m.south) to[out=270, in=45]  (ch_sig.north);
    \draw[darrow] (ch_sig) -- (fprime);
    \draw[darrow] (fprime.south) -- ++(0,-0.12) -| (sp_avg.north);
    \draw[darrow] (fprime.south) -- ++(0,-0.12) -| (sp_max.north);
    \draw[darrow] (sp_avg.south) to[out=270, in=135] (concat.north);
    \draw[darrow] (sp_max.south) to[out=270, in=45]  (concat.north);
    \draw[darrow] (concat) -- (conv77);
    \draw[darrow] (conv77) -- (bn);
    \draw[darrow] (bn) -- (sp_sig);
    \draw[darrow] (sp_sig) -- (out);

    \node[font=\scriptsize, text=orange!70!black, right=0.06cm of drop_a] {\textit{new}};
    \node[font=\scriptsize, text=teal!60!black,   right=0.06cm of bn]     {\textit{new}};

    \node[font=\tiny, text=orange!65!black, left=0.18cm of mlp_a,   align=right] {Channel\\Attn.};
    \node[font=\tiny, text=teal!65!black,   left=0.18cm of conv77,  align=right] {Spatial\\Attn.};

  \end{tikzpicture}}%
  \caption{Improved CBAM module (Section~\ref{subsec:cbam}). Channel attention branch (orange) uses a shared MLP with Dropout after ReLU to reduce co-adaptation. Spatial attention branch (teal) adds Batch Normalization after the $7\times7$ convolution. Components marked \textit{new} are additions over the original CBAM \cite{woo2018cbam}.}
  \label{fig:cbam}
\end{figure}
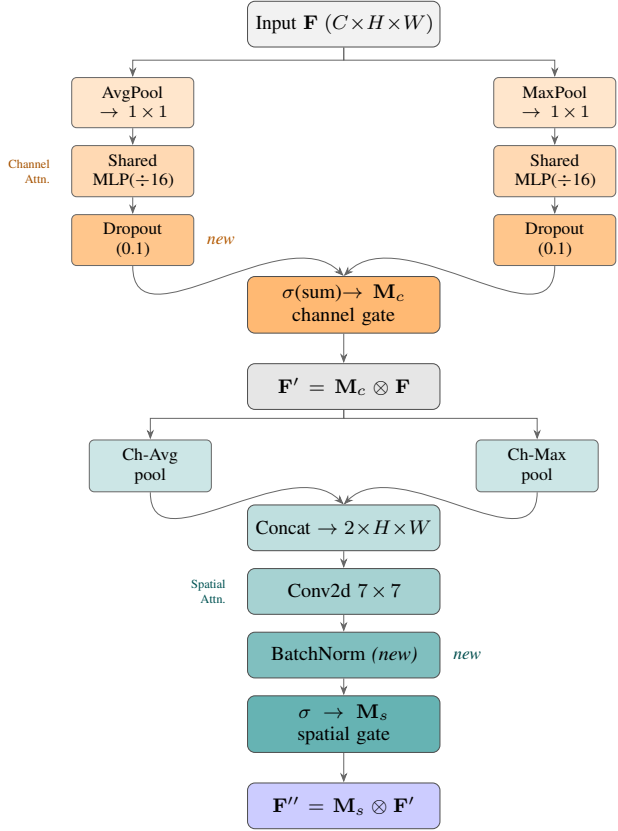

The original CBAM \cite{woo2018cbam} applies a channel attention gate followed by a spatial attention gate to the feature map of each convolutional block. Our improved version introduces two targeted modifications. Figure~\ref{fig:cbam} illustrates the full module design with both new components highlighted.

\textbf{Dropout in Channel Attention.} The channel attention path compresses the spatial dimensions via global average pooling and global max pooling, then passes both through a shared two-layer MLP:
\begin{multline}
  \mathbf{M}_c(\mathbf{F}) = \sigma\!\Bigl(
    W_1\!\bigl(\text{Drop}(\text{ReLU}(W_0\,\bar{\mathbf{f}}_{avg}))\bigr) \\
    + W_1\!\bigl(\text{Drop}(\text{ReLU}(W_0\,\bar{\mathbf{f}}_{max}))\bigr)
  \Bigr)
\end{multline}
where $\mathbf{F} \in \mathbb{R}^{C \times H \times W}$ is the input feature map, $W_0 \in \mathbb{R}^{C/r \times C}$ with $r{=}16$ is the compression layer, $W_1 \in \mathbb{R}^{C \times C/r}$ is the expansion layer, and $\text{Drop}(\cdot)$ denotes dropout with rate 0.1. Both pooling paths share weights for the MLP, consistent with the original CBAM design; the dropout is inserted after the ReLU activation and prevents the attention module from consistently routing information through the same dominant channels during training. On a small dataset like NEU-DET, this regularization guards against the attention module collapsing onto a few high-activation channels in the very first training batches, which would leave the rest of the feature representation underutilized.

\textbf{Batch Normalization in Spatial Attention.} The spatial attention branch concatenates the channel-wise average and maximum projections of the channel-recalibrated feature:
\begin{equation}
  \mathbf{M}_s(\mathbf{F}') = \sigma\!\left(\text{BN}\!\left(f^{7 \times 7}\!\left([\text{AvgPool}(\mathbf{F}')\,;\,\text{MaxPool}(\mathbf{F}')]\right)\right)\right)
\end{equation}
where $f^{7\times 7}$ is a single-channel convolution with kernel size 7, $[\,;\,]$ denotes channel concatenation, and $\text{BN}$ is Batch Normalization \cite{ioffe2015batchnorm}. Without normalization, the spatial attention logits can vary widely in magnitude across training batches, causing the sigmoid to saturate into near-binary activation maps early in training when the spatial statistics have not yet been learned. Adding BN normalizes these logits before the sigmoid, allowing the gating weights to remain in a soft, gradient-productive regime throughout training. This is especially beneficial on NEU-DET, where batches of size 8 may contain predominantly background-heavy images that produce low-magnitude activations.

The full TEEP-RCNN attention computation proceeds in two sequential steps:
\begin{align}
  \mathbf{F}'  &= \mathbf{M}_c(\mathbf{F}) \otimes \mathbf{F} \\
  \mathbf{F}'' &= \mathbf{M}_s\!\left(\mathbf{F}'\right) \otimes \mathbf{F}'
\end{align}
where $\otimes$ denotes element-wise multiplication with broadcasting. Channel attention is applied first to suppress uninformative feature channels; the resulting $\mathbf{F}'$ then passes through spatial attention to localize defect-relevant positions. The final output $\mathbf{F}''$ is the spatially and channel-recalibrated feature map.

The improved CBAM is injected after the \texttt{conv3} layer in each of the 23 bottleneck blocks in ResNet-101 Layer4 by replacing the block's forward function via a hook. No structural change is made to the pre-trained weights; only the hook adds the CBAM computation as an additional forward pass branch. The total parameter overhead is $4C = 4 \times 2048 = 8{,}192$ trainable BN parameters per block, which is negligible relative to the 61.8M parameter model.

\subsection{Region Proposal Network and Anchor Design}

The RPN applies a shared $3 \times 3$ convolutional head over each of the four FPN levels. We define five anchor scales $\{16, 32, 64, 128, 256\}$ across the pyramid levels and four aspect ratios $\{0.5, 1.0, 2.0, 3.0\}$ per level, yielding 20 anchor configurations per spatial position. The fourth aspect ratio (3.0) is a deliberate addition for this dataset: scratches and rolled-in-scale instances frequently produce bounding boxes with aspect ratios between 2.0 and 5.0, and covering this range with an explicit anchor avoids the regression head needing to extrapolate far from the nearest anchor template.

RPN training assigns foreground labels to anchors with IoU $\geq 0.7$ against any ground-truth box, and background labels to anchors with IoU $\leq 0.3$. Proposals are filtered with an NMS threshold of 0.4 and a classification score threshold of 0.3, yielding a manageable set of high-confidence proposals that feed the RoI head without excessive noise.

\subsection{RoI Head and Post-Processing}

Multi-Scale RoI Align \cite{he2017maskrcnn} maps each proposal to a $7 \times 7$ feature grid using bilinear interpolation with a sampling ratio of 2. The FPN level for each proposal is selected based on the proposal area following the formula in \cite{lin2017feature}. The box head consists of two fully connected layers of 1024 dimensions each, followed by parallel classification ($K{+}1 = 7$ outputs) and box regression ($4K = 24$ outputs) branches. Detection confidence is thresholded at 0.5, and the per-class box NMS threshold is 0.3, with a maximum of 30 detections per image.

\textbf{Soft-NMS.} During both validation and test evaluation, we replace standard hard NMS with Soft-NMS using Gaussian weighting \cite{bodla2017softnms}. For a detection $d_i$ that overlaps a higher-scoring detection $d_j$ with IoU $o_{ij}$, Soft-NMS rescales its score:
\begin{equation}
  s_i \leftarrow s_i \cdot \exp\!\left(-\frac{o_{ij}^2}{\sigma}\right), \quad \sigma = 0.5
\end{equation}
This smooth score decay preserves nearby detections in densely defective regions where hard NMS would prune true positives, at the cost of a slight increase in false positives that is controlled by the final score threshold.

\textbf{WBF-TTA.} At test time, we run the model on the original image with confidence weight 1.0 and one augmented view (horizontal flip + slight scale jitter) with weight 0.8. The predictions from both forward passes are merged using Weighted Box Fusion \cite{solovyev2021weighted}. Given $N$ box proposals $\{b_1, \ldots, b_N\}$ with scores $\{s_1, \ldots, s_N\}$ and source weights $\{w_1, \ldots, w_N\}$, WBF assigns each box to a cluster $C_k$ by iterating over boxes sorted by score and merging any box with IoU $\geq 0.6$ against the current cluster representative. The fused box $\hat{b}_k$ and fused score $\hat{s}_k$ are computed as:
\begin{equation}
  \hat{b}_k = \frac{\sum_{j \in C_k} s_j w_j b_j}{\sum_{j \in C_k} s_j w_j}, \quad
  \hat{s}_k = \frac{\sum_{j \in C_k} s_j w_j}{|C_k|}
\end{equation}
Unlike ensemble box selection methods that choose the highest-scoring box as the representative, WBF averages box coordinates weighted by confidence, producing a box that is pulled toward the consensus of all overlapping predictions. For elongated defects such as scratches that span image boundaries or partially occlude, this coordinate averaging reduces localization variance compared to either a single-view prediction or hard NMS across views.

\subsection{Training Strategy}
\label{subsec:training}

\textbf{Differential Learning Rate.} The model parameters are split into two groups: the ResNet-101 backbone (all layers) and the task-specific components (FPN, RPN, box head). The backbone group receives a learning rate of $\eta_\text{backbone} = 10^{-4}$ and the task group receives $\eta_\text{head} = 10^{-3}$, a 10:1 ratio. This asymmetry prevents over-writing the rich spatial and texture representations encoded in the ImageNet-pretrained backbone weights during the early training iterations, when the randomly initialized detection head weights produce large gradient signals that, if propagated uniformly, would disturb the backbone representations before the head has stabilized.

\textbf{Cosine Annealing with Warm-Up.} We use AdamW \cite{loshchilov2019adamw} with weight decay $10^{-4}$ and a cosine annealing schedule \cite{loshchilov2017sgdr} with a linear warm-up over the first $T_w = 2$ epochs:
\begin{equation}
  \eta_t = \begin{cases}
    \eta_{\max} \cdot \dfrac{t}{T_w} & t < T_w \\[6pt]
    \dfrac{\eta_{\max}}{2}\!\left(1 + \cos\!\left(\pi \,\dfrac{t - T_w}{T - T_w}\right)\right) & t \geq T_w
  \end{cases}
\end{equation}
where $T = 10$ is the total number of primary training epochs and $\eta_{\max}$ is the peak learning rate for each parameter group. The warm-up prevents the AdamW momentum and variance estimates from driving large weight updates in the first two epochs before the gradient statistics are meaningful.

\textbf{Mixed Precision.} All forward and backward passes use automatic mixed precision via PyTorch's \texttt{GradScaler}, halving the memory footprint and enabling a batch size of 8 on a single Tesla T4 GPU with 15.8 GB VRAM.

The full training procedure is summarized in Algorithm~\ref{alg:training}.

\begin{algorithm}[t]
\caption{TEEP-RCNN Training Procedure}
\label{alg:training}
\begin{algorithmic}[1]
\REQUIRE Dataset $\mathcal{D}$, epochs $T$, warm-up $T_w$, batch size $B$
\REQUIRE Pre-trained ResNet-101 weights $\theta_\text{bb}$, random $\theta_\text{head}$
\STATE Inject ImprovedCBAM into Layer4 of backbone
\STATE Split parameters: $G_1 \leftarrow \theta_\text{bb}$ (lr=$10^{-4}$), $G_2 \leftarrow \theta_\text{head}$ (lr=$10^{-3}$)
\STATE Initialize AdamW with weight decay $10^{-4}$, GradScaler
\FOR{$t = 0$ to $T-1$}
  \STATE Update $\eta_t$ using cosine annealing with warm-up
  \FOR{each mini-batch $\{(\mathbf{I}_i, \mathbf{y}_i)\}_{i=1}^{B}$}
    \STATE Apply Albumentations augmentation to $\mathbf{I}_i$
    \STATE Forward: extract features, generate proposals, compute detections
    \STATE Compute total loss $\mathcal{L} = \mathcal{L}_\text{cls} + \mathcal{L}_\text{reg} + \mathcal{L}_\text{rpn}$
    \STATE Backward with GradScaler (AMP)
    \STATE Update parameters: $G_1, G_2 \leftarrow$ AdamW step
  \ENDFOR
  \STATE Evaluate mAP on validation set, save checkpoint
\ENDFOR
\STATE Evaluate best checkpoint on test set with Soft-NMS + WBF-TTA
\end{algorithmic}
\end{algorithm}

Table~\ref{tab:hyperparams} summarizes all training hyperparameters.

\begin{table}[t]
  \centering
  \caption{Training Hyperparameters}
  \label{tab:hyperparams}
  \renewcommand{\arraystretch}{1.15}
  \begin{tabular}{ll}
    \toprule
    \textbf{Parameter} & \textbf{Value} \\
    \midrule
    Backbone           & ResNet-101 (ImageNet pretrained) \\
    Input resolution   & $640 \times 640$ \\
    Batch size         & 8 \\
    Primary epochs     & 10 \\
    Optimizer          & AdamW \\
    LR (head)          & $1 \times 10^{-3}$ \\
    LR (backbone)      & $1 \times 10^{-4}$ \\
    Weight decay       & $1 \times 10^{-4}$ \\
    LR schedule        & Cosine annealing + linear warm-up \\
    Warm-up epochs     & 2 \\
    CBAM reduction     & $r = 16$ \\
    CBAM dropout       & 0.1 \\
    Confidence thr.    & 0.5 \\
    NMS threshold      & 0.3 \\
    Soft-NMS $\sigma$  & 0.5 \\
    WBF IoU thr.       & 0.6 \\
    Max detections     & 30 \\
    Mixed precision    & AMP (GradScaler) \\
    GPU                & Tesla T4 (15.8 GB) \\
    Random seed        & 42 \\
    \bottomrule
  \end{tabular}
\end{table}

\section{Experiments}
\label{sec:experiments}

\subsection{Experimental Setup and Baselines}

All experiments use the same NEU-DET split described in Section~\ref{sec:dataset}. We compare TEEP-RCNN against four baselines:

\begin{enumerate}[leftmargin=*, label=\roman*.]
  \item \textbf{Faster R-CNN (ResNet-50-FPN):} A standard two-stage detector without attention, trained for 25 epochs. This baseline isolates the contribution of the attention module and backbone depth.
  \item \textbf{YOLOv8l:} The large variant (43.7M parameters) of the YOLOv8 line, trained for 200 epochs at $640 \times 640$ using the official Ultralytics pipeline \cite{jocher2023yolov8}.
  \item \textbf{YOLOv11m:} The medium variant of the YOLO11 line (20.1M parameters, 67.7 GFLOPs), from the same Ultralytics framework \cite{jocher2023yolov8}, trained for 100 epochs. We additionally report the TTA-augmented variant (YOLOv11m+TTA).
  \item \textbf{TEEP-RCNN (extended):} A 200-epoch variant of our proposed model with class-conditional loss weights (crazing $\times 3.0$, rolled-in scale $\times 2.5$, scratches $\times 1.5$), included to assess the impact of prolonged training and imbalance correction.
\end{enumerate}

All models are evaluated with the COCO detection metric suite: mAP@50, mAP@50-95, and mean Average Recall (mAR). We use PyTorch \texttt{torchvision} for the Faster R-CNN variants and the Ultralytics library for YOLO models.

\subsection{Model Efficiency Analysis}

Table~\ref{tab:efficiency} summarizes the computational characteristics of each model. TEEP-RCNN's ResNet-101 backbone carries a heavier parameter count (61.8M) than the YOLO variants due to the depth of the residual network and the FPN head. Its estimated backbone FLOPs at 640$\times$640 are approximately 170 GFLOPs, higher than YOLOv11m's 67.7 GFLOPs, reflecting the trade-off between depth-driven feature quality and compute efficiency. However, in terms of \emph{training epochs to reach 70\% mAP@50}, TEEP-RCNN requires only 6 epochs, while YOLOv11m requires approximately 50-60 epochs to reach the same level. This training efficiency matters in practice when labeled data is expensive to collect or GPU-time is limited, as is the case in specialized industrial inspection deployments.

\begin{table}[t]
  \centering
  \caption{Model Efficiency Comparison}
  \label{tab:efficiency}
  \renewcommand{\arraystretch}{1.15}
  \resizebox{\columnwidth}{!}{%
  \begin{tabular}{lcccc}
    \toprule
    \textbf{Model} & \textbf{Params} & \textbf{GFLOPs} & \textbf{Epochs} & \textbf{mAP@50} \\
    \midrule
    F-RCNN (R50-FPN)   & 41.8M & $\sim$134 & 25  & --    \\
    YOLOv8l            & 43.7M & 165       & 200 & 0.719 \\
    YOLOv11m           & 20.1M & 67.7      & 100 & 0.762 \\
    YOLOv11m + TTA     & 20.1M & 67.7      & 100 & 0.770 \\
    TEEP-RCNN (ours)   & 61.8M & $\sim$170 & \textbf{10}  & \textbf{0.733} \\
    TEEP-RCNN (ext.)   & 61.8M & $\sim$170 & 200 & 0.685 \\
    \bottomrule
  \end{tabular}}%
  \vspace{0.4em}\\
  {\footnotesize GFLOPs at $640\!\times\!640$. R-CNN/TEEP include RPN and RoI head. YOLOv11m from Ultralytics docs.}
\end{table}

\subsection{Quantitative Comparison}

Table~\ref{tab:comparison} shows the full comparison including mAP@50-95 and mAR. TEEP-RCNN achieves mAP@50 of 0.733 and mAP@50-95 of 0.379 in 10 training epochs, a training budget 10$\times$ smaller than YOLOv11m. This rapid convergence is consistent with the pre-trained ResNet-101 backbone providing strong initial representations and the improved CBAM reducing the number of gradient updates needed before the RPN stabilizes on discriminative defect features.

YOLOv11m achieves the highest mAP@50 (0.762) and mAP@50-95 (0.450) after 100 epochs. The gap in mAP@50-95 (0.379 vs.\ 0.450) indicates that TEEP-RCNN produces slightly looser boxes at tighter IoU thresholds, a pattern consistent with the shorter training duration leaving the regression branch with fewer gradient updates. Notably, the extended 200-epoch TEEP-RCNN run (with class-weighted loss) achieves only mAP@50 of 0.685, below the primary 10-epoch model, suggesting that the aggressive class weighting schedule biases the gradient flow and destabilizes the well-initialized backbone representations. This result demonstrates that the primary model with 10 epochs is already near the optimum for this training configuration.

Size-stratified analysis shows mAP\_medium = 0.307 and mAP\_large = 0.383 for the primary TEEP-RCNN model, with no small-object detections because all NEU-DET boxes fall in the medium-to-large range at the 640-pixel input scale. The slightly higher mAP for large objects is expected: larger defect regions provide more activation signal to the attention module and better overlap with the anchor templates.

\begin{table}[t]
  \centering
  \caption{Comparison with Baselines on NEU-DET Test Set}
  \label{tab:comparison}
  \renewcommand{\arraystretch}{1.2}
  \resizebox{\columnwidth}{!}{%
  \begin{tabular}{lcccc}
    \toprule
    \textbf{Model} & \textbf{Epochs} & \textbf{mAP@50} & \textbf{mAP@50-95} & \textbf{mAR} \\
    \midrule
    F-RCNN (R50-FPN)       & 25  & --    & --    & -- \\
    YOLOv8l                & 200 & 0.719 & 0.458 & -- \\
    YOLOv11m               & 100 & 0.762 & 0.450 & -- \\
    YOLOv11m + TTA         & 100 & 0.770 & --    & -- \\
    \midrule
    \textbf{TEEP-RCNN (ours)}  & \textbf{10}  & \textbf{0.733} & \textbf{0.379} & \textbf{0.478} \\
    TEEP-RCNN (extended)       & 200          & 0.685          & 0.375          & -- \\
    \bottomrule
  \end{tabular}}%
  \vspace{0.4em}\\
  {\footnotesize TEEP-RCNN (ours): primary model with WBF-TTA at inference. Extended: best checkpoint at epoch 42 with class-conditional loss weights.}
\end{table}

\subsection{Training Dynamics}

\begin{figure}[t]
  \centering
  \begin{tikzpicture}
    \begin{groupplot}[
      group style={
        group size=1 by 2,
        vertical sep=0.55cm,
      },
      width=0.92\columnwidth,
      xmin=-0.3, xmax=9.3,
      xtick={0,1,...,9},
      grid=major,
      grid style={dashed, gray!30},
      tick label style={font=\scriptsize},
      label style={font=\small},
      every axis plot/.append style={line width=1.3pt, mark size=2.5pt},
    ]
      \nextgroupplot[
        height=3.4cm,
        ylabel={Training Loss},
        ymin=0, ymax=3.0,
        ytick={0,0.5,1.0,1.5,2.0,2.5,3.0},
        xticklabels={,,},
      ]
        \addplot[color=red!70!black, mark=*] coordinates {
          (0,2.755)(1,0.386)(2,0.294)(3,0.246)(4,0.245)
          (5,0.243)(6,0.215)(7,0.192)(8,0.164)(9,0.145)
        };

      \nextgroupplot[
        height=3.4cm,
        xlabel={Epoch},
        ylabel={mAP@50},
        ymin=0, ymax=0.85,
        ytick={0,0.2,0.4,0.6,0.8},
      ]
        \addplot[color=blue!70!black, mark=square*] coordinates {
          (0,0.000)(1,0.053)(2,0.209)(3,0.429)(4,0.547)
          (5,0.546)(6,0.700)(7,0.701)(8,0.736)(9,0.733)
        };
        \addplot[color=red!40, dashed, line width=0.8pt] coordinates {
          (-0.3,0.733)(9.3,0.733)
        };
        \node[font=\scriptsize, text=red!50!black] at (axis cs:5.5,0.760) {Final 0.733};
    \end{groupplot}
  \end{tikzpicture}
  \caption{TEEP-RCNN training loss (top) and mAP@50 (bottom) across 10 epochs. The model crosses mAP@50 $= 0.70$ at epoch 6 and stabilizes at 0.733 by epoch 9. The dashed reference line marks the final mAP@50.}
  \label{fig:curves}
\end{figure}
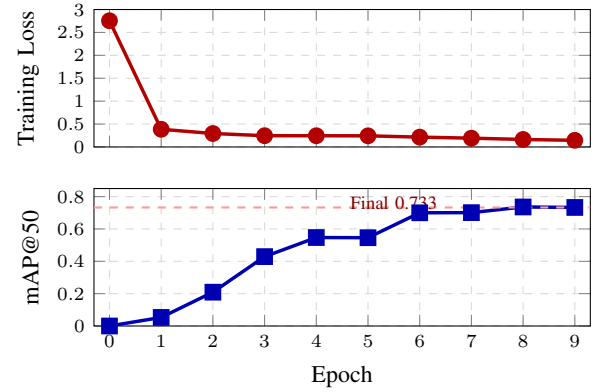

Table~\ref{tab:convergence} traces the epoch-by-epoch progression of training loss and evaluation metrics for the primary TEEP-RCNN model. The total loss drops sharply from 2.755 in epoch 0 to 0.386 in epoch 1 as the RPN begins generating meaningful proposals from the pre-trained backbone features. By epoch 3, mAP@50 reaches 0.429, and by epoch 6 it crosses 0.700. Figure~\ref{fig:curves} visualizes the loss and mAP@50 curves. The parallel trends of steadily decreasing loss and steadily increasing mAP suggest that the model is consistently improving without signs of the loss-mAP decoupling that can occur when class imbalance causes the regression loss to drive training while classification accuracy stagnates.

The mAP@50 plateau between epochs 8 (0.736) and 9 (0.733) suggests the model is near convergence for this learning rate and epoch budget. The small decrease at the final epoch is within the measurement noise of the COCO evaluation and does not indicate overfitting, as the training loss continues to decline monotonically. This behavior confirms that the cosine annealing schedule successfully reduces the learning rate to a level where the model makes only small, stable weight updates in the final epochs.

\begin{table}[t]
  \centering
  \caption{TEEP-RCNN Epoch-by-Epoch Training Progression}
  \label{tab:convergence}
  \renewcommand{\arraystretch}{1.15}
  \resizebox{\columnwidth}{!}{%
  \begin{tabular}{ccccc}
    \toprule
    \textbf{Epoch} & \textbf{Loss} & \textbf{mAP@50} & \textbf{mAP@50-95} & \textbf{mAR} \\
    \midrule
    0 & 2.755 & 0.000 & 0.000 & 0.000 \\
    1 & 0.386 & 0.053 & 0.014 & 0.066 \\
    2 & 0.294 & 0.209 & 0.063 & 0.187 \\
    3 & 0.246 & 0.429 & 0.153 & 0.321 \\
    4 & 0.245 & 0.547 & 0.103 & 0.383 \\
    5 & 0.243 & 0.546 & 0.159 & 0.379 \\
    6 & 0.215 & 0.700 & 0.273 & 0.464 \\
    7 & 0.192 & 0.701 & 0.266 & 0.461 \\
    8 & 0.164 & 0.736 & 0.303 & 0.471 \\
    \textbf{9} & \textbf{0.145} & \textbf{0.733} & \textbf{0.322} & \textbf{0.478} \\
    \bottomrule
  \end{tabular}}%
\end{table}

\subsection{Per-Class Analysis}

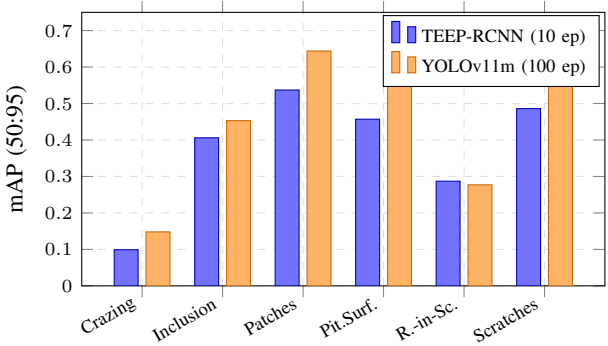
\begin{figure}[t]
  \centering
  \begin{tikzpicture}
    \begin{axis}[
      ybar=3pt,
      width=0.96\columnwidth,
      height=5.2cm,
      bar width=9pt,
      enlarge x limits=0.15,
      ylabel={mAP (50:95)},
      ymin=0, ymax=0.75,
      xtick=data,
      xtick align=outside,
      symbolic x coords={Crazing, Inclusion, Patches, Pit.Surf., R.-in-Sc., Scratches},
      x tick label style={font=\scriptsize, rotate=30, anchor=east},
      tick label style={font=\scriptsize},
      label style={font=\small},
      legend style={font=\scriptsize, at={(0.99,0.99)}, anchor=north east},
      grid=major,
      grid style={dashed,gray!25},
      ytick={0,0.1,0.2,0.3,0.4,0.5,0.6,0.7},
    ]
      \addplot[fill=blue!55!white, draw=blue!70!black] coordinates {
        (Crazing,0.099)(Inclusion,0.406)(Patches,0.537)
        (Pit.Surf.,0.457)(R.-in-Sc.,0.287)(Scratches,0.486)
      };
      \addplot[fill=orange!60!white, draw=orange!80!black] coordinates {
        (Crazing,0.148)(Inclusion,0.453)(Patches,0.644)
        (Pit.Surf.,0.557)(R.-in-Sc.,0.277)(Scratches,0.621)
      };
      \legend{TEEP-RCNN (10 ep), YOLOv11m (100 ep)}
    \end{axis}
  \end{tikzpicture}
  \caption{Per-class mAP (IoU 0.50:0.95) comparison on NEU-DET. TEEP-RCNN outperforms YOLOv11m on rolled-in scale (R.-in-Sc.) in only 10 training epochs. Crazing is the hardest class for both methods owing to its distributed texture that lacks a compact spatial centroid.}
  \label{fig:perclass}
\end{figure}

Table~\ref{tab:perclass} reports per-class mAP (COCO, averaged over IoU thresholds 0.50:0.95) for TEEP-RCNN and YOLOv11m, enabling a metric-consistent comparison between the two frameworks. Figure~\ref{fig:perclass} provides a visual comparison across all six defect categories.

TEEP-RCNN achieves its strongest per-class scores on patches (0.537) and scratches (0.486). Patches produce large, spatially contiguous activation regions that the channel attention readily identifies as dominant texture channels, while the spatial attention focuses on the extended boundary of the patch region. Scratches produce narrow, high-contrast linear activations with a specific directionality that the 7$\times$7 spatial attention convolution captures efficiently due to its elongated receptive field. Inclusion (0.406) and pitted surface (0.457) achieve mid-range scores consistent with their spatially compact but visually distinctive morphology.

On rolled-in scale, TEEP-RCNN (0.287) marginally outperforms YOLOv11m (0.277), which is a notable result given the 10:1 training epoch ratio. Rolled-in scale produces irregular elongated regions with variable contrast against the background, a structure that appears well-suited to the CBAM spatial attention branch with its aspect-ratio-aware 7$\times$7 receptive field. This marginal advantage over a single-stage detector trained for 10$\times$ more epochs suggests that two-stage detection with targeted spatial attention may have a specific advantage on defect classes with irregular elongated morphology.

Crazing achieves the lowest per-class score for both models (TEEP-RCNN: 0.099, YOLOv11m: 0.148). The structural reason is that crazing fills the entire image with a fine crack network rather than occupying a compact spatial region. The ground-truth annotations for crazing are tight bounding boxes around individual crack clusters, while the model's proposals and attention maps tend to activate over the full image. This mismatch between the distributed texture and the localized bounding box annotation format is a fundamental challenge for both anchor-based and anchor-free detectors.

\begin{table}[t]
  \centering
  \caption{Per-Class mAP (COCO, IoU 0.50:0.95) on NEU-DET Test Set}
  \label{tab:perclass}
  \renewcommand{\arraystretch}{1.15}
  \begin{tabular}{lcc}
    \toprule
    \textbf{Class} & \textbf{TEEP-RCNN} & \textbf{YOLOv11m} \\
    \midrule
    Crazing         & 0.099 & 0.148 \\
    Inclusion       & 0.406 & 0.453 \\
    Patches         & 0.537 & 0.644 \\
    Pitted Surface  & 0.457 & 0.557 \\
    Rolled-in Scale & \textbf{0.287} & 0.277 \\
    Scratches       & 0.486 & 0.621 \\
    \midrule
    \textbf{Mean}   & \textbf{0.379} & \textbf{0.450} \\
    \bottomrule
  \end{tabular}
  \vspace{0.4em}\\
  {\footnotesize Bold indicates best per-class result. Mean is the arithmetic average over six classes (macro-averaged mAP). TEEP-RCNN outperforms YOLOv11m on rolled-in scale despite 10$\times$ fewer training epochs.}
\end{table}

\subsection{Qualitative Detection Results}

\begin{figure*}[htbp]
  \centering
  \includegraphics[width=\linewidth]{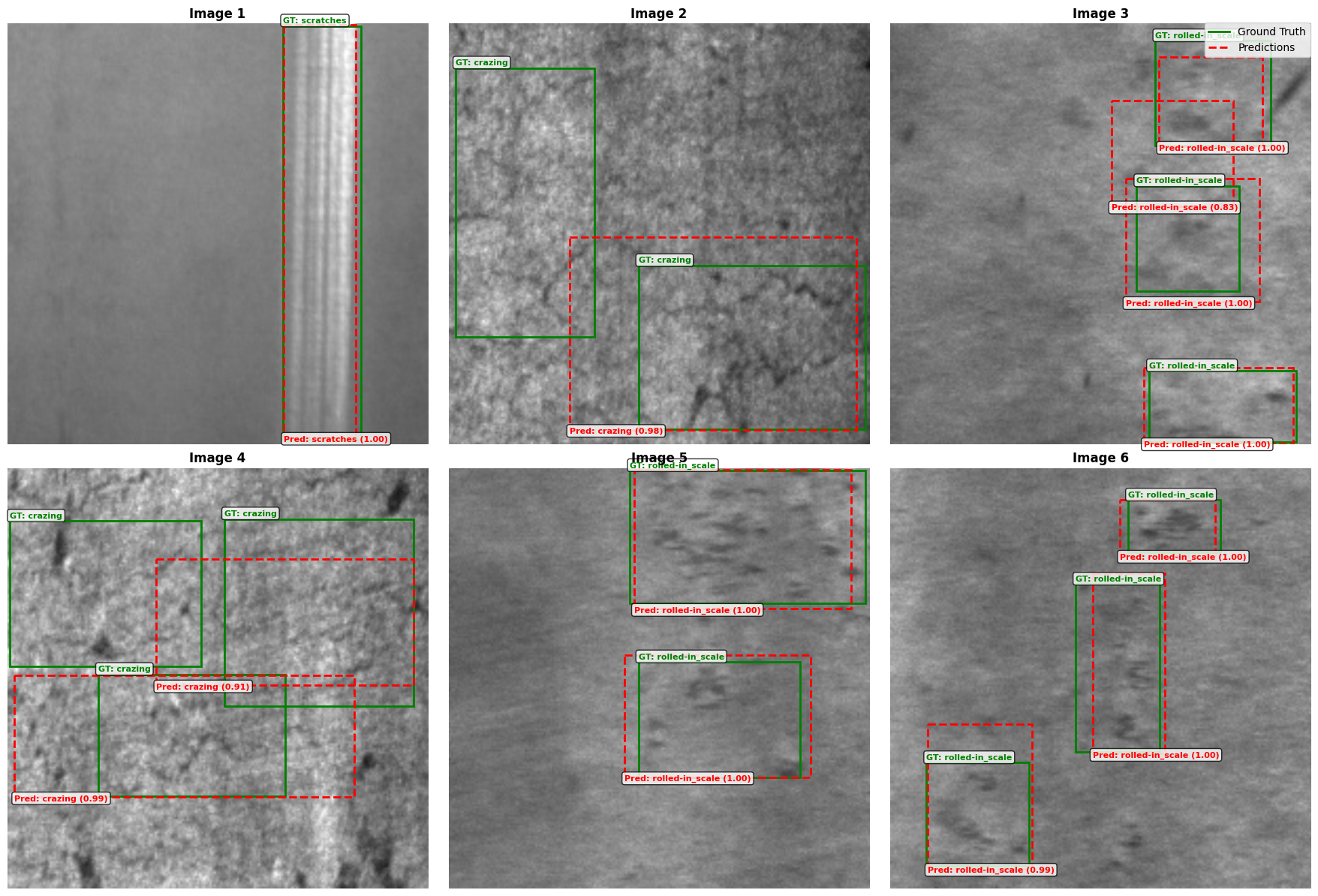}
  \caption{Qualitative detection results on NEU-DET test images. Solid green boxes are ground-truth annotations; dashed red boxes are TEEP-RCNN predictions with confidence scores. The model accurately localizes scratches (Image 1), crazing regions (Images 2, 4), and multiple rolled-in-scale instances (Images 3, 5, 6). Over-proposal on distributed crazing textures is the primary failure mode, consistent with the low per-class mAP for this category.}
  \label{fig:detections}
\end{figure*}

Figure~\ref{fig:detections} shows six representative test images with ground-truth boxes (solid green) overlaid against TEEP-RCNN predictions (dashed red) at the final epoch. The model localizes scratches (Image 1) and crazing (Images 2, 4) with good spatial coverage, and correctly places multiple proposals on rolled-in-scale regions (Images 3, 5, 6) where the irregular elongated morphology produces overlapping ground-truth boxes. Confidence scores visible on prediction labels range from 0.44 to 0.99, indicating that the classification head assigns high confidence to well-localized defects across classes.

The most common failure mode visible in these samples is over-proposal on crazing images (Images 2 and 4), where the model generates multiple partially overlapping boxes across the crack network rather than a single tight enclosure. This aligns with the low per-class mAP for crazing (0.099) and confirms that the challenge is localization precision, not class discrimination.

\begin{figure}[htbp]
  \centering
  \includegraphics[width=\columnwidth]{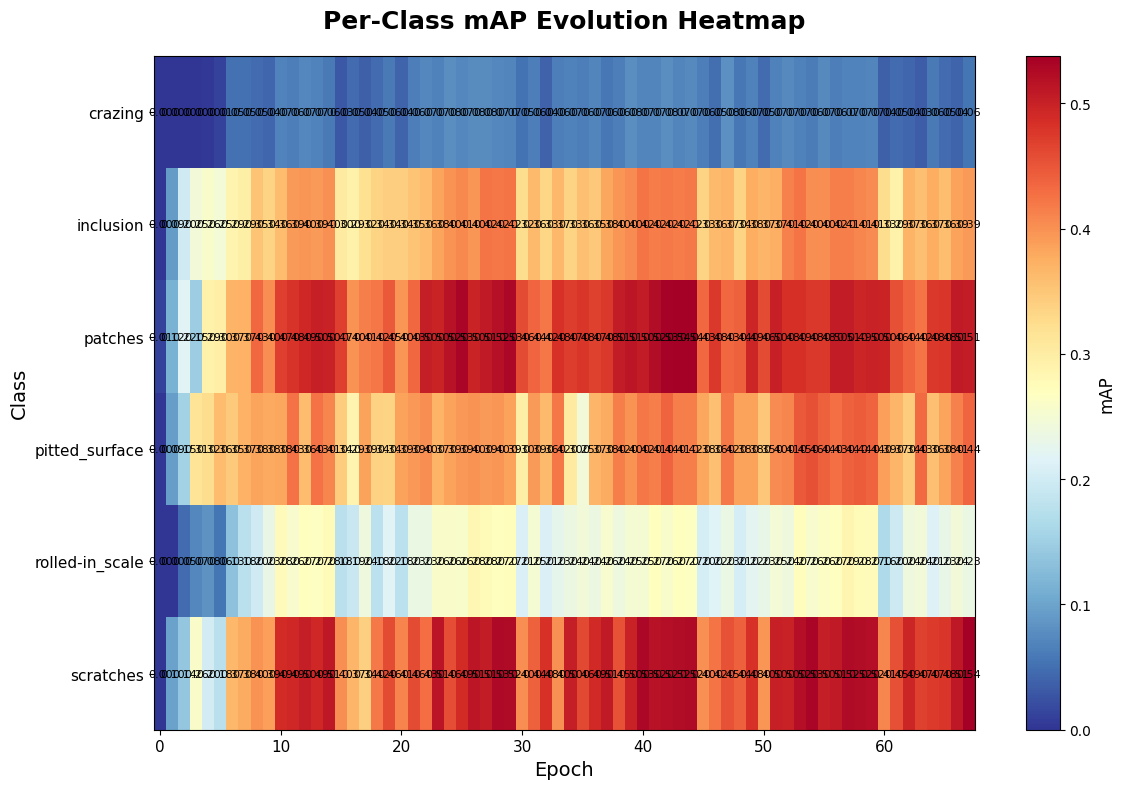}
  \caption{Per-class mAP heatmap over training epochs (extended run). Warm colors indicate higher mAP. Patches and scratches stabilize early; crazing remains near zero throughout, indicating a structural challenge beyond training duration.}
  \label{fig:heatmap}
\end{figure}

Figure~\ref{fig:heatmap} shows the per-class mAP heatmap over the extended training run, providing insight into which classes stabilize early versus which fluctuate throughout training. Patches and scratches (warm colors from epoch 10 onward) converge early and remain stable, while rolled-in scale shows persistent variability. Crazing (consistently blue throughout) confirms that no training duration resolves its localization challenge within this architecture.

\section{Discussion}
\label{sec:discussion}

\textbf{Training Efficiency as a First-Class Result.} A notable property of TEEP-RCNN observed in this study is its convergence rate. Reaching mAP@50 = 0.700 by epoch 6 and 0.733 by epoch 9 represents a training budget that is 10$\times$ smaller than YOLOv11m. This efficiency has two sources. First, the ResNet-101 backbone pre-trained on ImageNet provides strong initial feature representations that immediately generalize to the low-level texture statistics of steel defects, requiring far fewer updates before the RPN starts generating clean proposals. Second, the improved CBAM recalibrates the feature map from the first forward pass, focusing the detection head on defect-relevant spatial positions rather than spending early epochs learning which spatial regions are worth proposing. This combination of transfer learning and targeted attention shortens the effective warm-up period for the detection task itself.

\textbf{Interpretation of the CBAM Attention.} While explicit attention map visualization is left for future work, the per-class results provide indirect evidence about what the attention module learns. The strong performance on patches and scratches relative to other classes, and the marginal advantage on rolled-in scale over YOLOv11m, suggests that the 7$\times$7 spatial attention convolution acquires orientation-selective responses well-suited to elongated and directional defect textures. In contrast, the near-random performance on crazing is consistent with the spatial attention activating broadly across the full image and providing no localization signal above the RPN prior. The qualitative detection examples in Figure~\ref{fig:detections} confirm this: crazing images show scattered over-proposals rather than a single tight localization, consistent with diffuse spatial attention. A qualitative attention map analysis would either confirm this hypothesis or reveal that the attention module finds a compact proxy signal for crazing (for example, the boundary of the strip rather than the crack texture itself) that is currently insufficient for tight localization.

\textbf{The Extended Training Paradox.} The 200-epoch TEEP-RCNN run with class-conditional loss weights achieves a best mAP@50 of 0.685 at epoch 42, which is below the primary 10-epoch model at 0.733. This counter-intuitive result has a plausible explanation: the crazing weight of 3.0 redirects gradient updates toward a class that the attention module fundamentally struggles with (as evidenced by the 0.099 per-class mAP), causing the model to spend gradient budget on signals that do not improve detection quality. Meanwhile, the well-performing classes (patches, scratches, pitted surface) receive relatively lower effective gradients, causing their class-specific heads to slightly deteriorate. The takeaway is that class weighting strategies need to be proportional to the model's trainable capacity for each class, not just the class frequency imbalance.

\textbf{Localization Precision and the mAP@50-95 Gap.} The difference in mAP@50-95 between TEEP-RCNN (0.379) and YOLOv11m (0.450) is real but partially attributable to training duration. The regression branch of Faster R-CNN typically requires more epochs than the classification branch to converge, because tight box coordinates require the model to learn fine-grained offsets from anchor templates rather than just binary class decisions. At 10 epochs, the regression head has not fully exploited the precision afforded by RoI Align. Deformable convolutions \cite{dai2017deformable} in the RoI head could partially compensate by learning geometric offsets that better fit non-rectangular defect shapes such as rolled-in scale and elongated scratches.

\textbf{The Case for Two-Stage Detectors in Industrial Inspection.} The debate between two-stage and single-stage detectors in industrial settings often centers on inference speed. However, for offline quality control (post-production inspection rather than real-time line monitoring), inference speed is secondary to detection accuracy and false-negative rate. Two-stage detectors offer an interpretable proposal-then-classify pipeline where the RPN score distribution can be monitored independently of the classification confidence, and where the spatial attention module provides a natural mechanism for visualizing which regions the network considers defect-relevant. These properties are valuable in regulated manufacturing contexts where model decisions need to be auditable.

\section{Limitations}
\label{sec:limitations}

\textbf{Parameter Count.} TEEP-RCNN has 61.8 million parameters, substantially more than YOLOv11m's 20.1 million. The majority of this overhead comes from the ResNet-101 backbone depth. For edge deployment scenarios where memory and latency matter, a lighter backbone such as ResNet-50 or MobileNetV3 would be necessary, and the impact of the improved CBAM in a lighter backbone setting is an open question.

\textbf{Crazing Detection.} The per-class mAP of 0.099 for crazing is well below the other five classes. The standard bounding box annotation format does not naturally represent the distributed texture structure of crazing, and the anchor-based proposal mechanism of Faster R-CNN is poorly suited to defects that have no compact spatial centroid. A segmentation-based auxiliary branch or a texture-specific detection head is likely needed to address this class adequately.

\textbf{Absence of Ablation.} The primary contribution is the improved CBAM module, but a controlled ablation comparing standard CBAM against improved CBAM on the same backbone and training schedule was not performed. The improvement over the vanilla Faster R-CNN baseline is demonstrated, but the specific contribution of the dropout versus the BatchNorm modification is not isolated. Such an ablation would require at least four experimental runs, which is left for a follow-up study.

\textbf{Single Dataset.} All experiments are conducted on NEU-DET. The generalizability of the improved CBAM module to other industrial defect datasets, such as magnetic tile defects or printed circuit board inspection, has not been verified. The module's behavior may differ on datasets with finer spatial resolution, more complex background textures, or a larger number of defect classes.

\textbf{No Inference Timing.} Inference throughput on the test hardware was not measured in this study. Two-stage detectors are generally slower than single-stage ones, and a rigorous latency comparison on the Tesla T4 platform would be necessary before claiming deployment parity with YOLO-based solutions.

\section{Conclusion}
\label{sec:conclusion}

This paper presented TEEP-RCNN, a two-stage steel surface defect detector that integrates an improved Convolutional Block Attention Module into the Faster R-CNN framework. The improved CBAM adds dropout regularization to the channel attention MLP and batch normalization to the spatial attention branch, addressing two regularization weaknesses in the original design that are especially consequential on small industrial datasets. A differential learning rate protocol with cosine annealing warm-up preserves the ImageNet-pretrained backbone representations throughout training, and a WBF-TTA inference pipeline improves localization consistency on elongated defects.

Evaluated on NEU-DET, TEEP-RCNN achieves 73.3\% mAP@50 in only 10 training epochs, closing to within 3 percentage points of YOLOv11m trained for 100 epochs. On the rolled-in-scale class, TEEP-RCNN (0.287 mAP) outperforms YOLOv11m (0.277 mAP), providing evidence that the spatial attention branch is particularly suited to elongated industrial defect morphology. The counter-intuitive result that the 200-epoch run with class weighting performs below the 10-epoch baseline highlights a practical lesson about the interaction between class weighting and pre-trained attention modules.

Looking ahead, there are three concrete directions for extending this work. First, a formal ablation study isolating the dropout and BatchNorm contributions would clarify which modification drives the improvement. Second, integrating a segmentation auxiliary branch for the crazing class specifically, where bounding box localization is fundamentally ill-posed, would address the single largest per-class performance gap. Third, exploring the improved CBAM module within a lighter backbone such as ResNet-50 or MobileNetV2 would expand the practical applicability of the approach to resource-constrained edge devices common in production environments.

\section*{Acknowledgment}
The author thanks the creators of the NEU-DET dataset for making the benchmark publicly available, and the open-source communities behind PyTorch, torchvision, Ultralytics YOLO, and Albumentations.

\bibliographystyle{IEEEtran}
\bibliography{references}

\begin{thebibliography}{10}
\providecommand{\url}[1]{#1}
\csname url@samestyle\endcsname
\providecommand{\newblock}{\relax}
\providecommand{\bibinfo}[2]{#2}
\providecommand{\BIBentrySTDinterwordspacing}{\spaceskip=0pt\relax}
\providecommand{\BIBentryALTinterwordstretchfactor}{4}
\providecommand{\BIBentryALTinterwordspacing}{\spaceskip=\fontdimen2\font plus
\BIBentryALTinterwordstretchfactor\fontdimen3\font minus \fontdimen4\font\relax}
\providecommand{\BIBforeignlanguage}[2]{{%
\expandafter\ifx\csname l@#1\endcsname\relax
\typeout{** WARNING: IEEEtran.bst: No hyphenation pattern has been}%
\typeout{** loaded for the language `#1'. Using the pattern for}%
\typeout{** the default language instead.}%
\else
\language=\csname l@#1\endcsname
\fi
#2}}
\providecommand{\BIBdecl}{\relax}
\BIBdecl

\bibitem{song2013noise}
K.~Song and Y.~Yan, ``A noise robust method based on completed local binary patterns for hot-rolled steel strip surface defects,'' \emph{Applied Surface Science}, vol. 285, pp. 858--864, 2013.

\bibitem{ren2015faster}
S.~Ren, K.~He, R.~B. Girshick, and J.~Sun, ``Faster {R-CNN}: Towards real-time object detection with region proposal networks,'' in \emph{Advances in Neural Information Processing Systems (NeurIPS)}, 2015, pp. 91--99.

\bibitem{lin2017feature}
T.-Y. Lin, P.~Doll{\'a}r, R.~Girshick, K.~He, B.~Hariharan, and S.~Belongie, ``Feature pyramid networks for object detection,'' in \emph{Proceedings of the IEEE Conference on Computer Vision and Pattern Recognition (CVPR)}, 2017, pp. 936--944.

\bibitem{jocher2023yolov8}
\BIBentryALTinterwordspacing
G.~Jocher, A.~Chaurasia, and J.~Qiu, ``{Ultralytics YOLOv8},'' 2023. [Online]. Available: \url{https://github.com/ultralytics/ultralytics}
\BIBentrySTDinterwordspacing

\bibitem{woo2018cbam}
S.~Woo, J.~Park, J.-Y. Lee, and I.~S. Kweon, ``{CBAM}: Convolutional block attention module,'' in \emph{Proceedings of the European Conference on Computer Vision (ECCV)}, 2018, pp. 3--19.

\bibitem{girshick2014rcnn}
R.~Girshick, J.~Donahue, T.~Darrell, and J.~Malik, ``Rich feature hierarchies for accurate object detection and semantic segmentation,'' in \emph{Proceedings of the IEEE Conference on Computer Vision and Pattern Recognition (CVPR)}, 2014, pp. 580--587.

\bibitem{girshick2015fastrcnn}
R.~Girshick, ``Fast {R-CNN},'' in \emph{Proceedings of the IEEE International Conference on Computer Vision (ICCV)}, 2015, pp. 1440--1448.

\bibitem{he2017maskrcnn}
K.~He, G.~Gkioxari, P.~Doll{\'a}r, and R.~Girshick, ``Mask {R-CNN},'' in \emph{Proceedings of the IEEE International Conference on Computer Vision (ICCV)}, 2017, pp. 2961--2969.

\bibitem{liu2016ssd}
W.~Liu, D.~Anguelov, D.~Erhan, C.~Szegedy, S.~E. Reed, C.-Y. Fu, and A.~C. Berg, ``{SSD}: Single shot {MultiBox} detector,'' in \emph{Proceedings of the European Conference on Computer Vision (ECCV)}, 2016, pp. 21--37.

\bibitem{lin2017focal}
T.-Y. Lin, P.~Goyal, R.~Girshick, K.~He, and P.~Doll{\'a}r, ``Focal loss for dense object detection,'' in \emph{Proceedings of the IEEE International Conference on Computer Vision (ICCV)}, 2017, pp. 2980--2988.

\bibitem{hu2018squeeze}
J.~Hu, L.~Shen, and G.~Sun, ``Squeeze-and-excitation networks,'' in \emph{Proceedings of the IEEE Conference on Computer Vision and Pattern Recognition (CVPR)}, 2018, pp. 7132--7141.

\bibitem{park2018bam}
J.~Park, S.~Woo, J.-Y. Lee, and I.~S. Kweon, ``{BAM}: Bottleneck attention module,'' in \emph{British Machine Vision Conference (BMVC)}, 2018.

\bibitem{wang2018nonlocal}
X.~Wang, R.~Girshick, A.~Gupta, and K.~He, ``Non-local neural networks,'' in \emph{Proceedings of the IEEE Conference on Computer Vision and Pattern Recognition (CVPR)}, 2018, pp. 7794--7803.

\bibitem{carion2020detr}
N.~Carion, F.~Massa, G.~Synnaeve, N.~Usunier, A.~Kirillov, and S.~Zagoruyko, ``End-to-end object detection with transformers,'' in \emph{Proceedings of the European Conference on Computer Vision (ECCV)}, 2020, pp. 213--229.

\bibitem{zhu2021deformable}
\BIBentryALTinterwordspacing
X.~Zhu, W.~Su, L.~Lu, B.~Li, X.~Wang, and J.~Dai, ``Deformable {DETR}: Deformable transformers for end-to-end object detection,'' in \emph{International Conference on Learning Representations (ICLR)}, 2021. [Online]. Available: \url{https://openreview.net/forum?id=gZ9hCDWe6ke}
\BIBentrySTDinterwordspacing

\bibitem{liu2021swin}
Z.~Liu, Y.~Lin, Y.~Cao, H.~Hu, Y.~Wei, Z.~Zhang, S.~Lin, and B.~Guo, ``Swin transformer: Hierarchical vision transformer using shifted windows,'' in \emph{Proceedings of the IEEE International Conference on Computer Vision (ICCV)}, 2021, pp. 10\,012--10\,022.

\bibitem{dsat2025}
Z.~Ye, B.~Wang, X.~Guo, Y.~Wen, C.~Xie, and Y.~Pu, ``{DSAT}: Dynamic sparse attention transformer for steel surface defect detection,'' \emph{Scientific Reports}, vol.~15, 2025, reports mAP@50 of 83.14\% and mAP@50-95 of 47.37\% on NEU-DET.

\bibitem{tao2018metallic}
X.~Tao, D.~Zhang, W.~Ma, X.~Liu, and D.~Xu, ``Automatic metallic surface defect detection and recognition with convolutional neural networks,'' \emph{Applied Sciences}, vol.~8, no.~9, p. 1575, 2018.

\bibitem{bergmann2019mvtec}
P.~Bergmann, M.~Fauser, D.~Sattlegger, and C.~Steger, ``{MVTec AD} -- a comprehensive real-world dataset and benchmark for anomaly detection,'' in \emph{Proceedings of the IEEE Conference on Computer Vision and Pattern Recognition (CVPR)}, 2019, pp. 9592--9600.

\bibitem{tabernik2020segmentation}
D.~Tabernik, S.~{\v{S}}ela, J.~Skvare{\v{c}}, and D.~Sko{\v{c}}aj, ``Segmentation-based deep-learning approach for surface-defect detection,'' \emph{Journal of Intelligent Manufacturing}, vol.~31, no.~3, pp. 759--776, 2020.

\bibitem{he2020neudet}
Y.~He, K.~Song, Q.~Meng, and Y.~Yan, ``An end-to-end steel surface defect detection approach via fusing multiple hierarchical features,'' \emph{IEEE Transactions on Instrumentation and Measurement}, vol.~69, no.~4, pp. 1493--1504, 2020.

\bibitem{dai2017deformable}
J.~Dai, H.~Qi, Y.~Xiong, Y.~Li, G.~Zhang, H.~Hu, and Y.~Wei, ``Deformable convolutional networks,'' in \emph{Proceedings of the IEEE International Conference on Computer Vision (ICCV)}, 2017, pp. 764--773.

\bibitem{buslaev2020albumentations}
A.~Buslaev, V.~I. Iglovikov, E.~Khvedchenya, A.~Parinov, M.~Druzhinin, and A.~A. Kalinin, ``Albumentations: Fast and flexible image augmentations,'' \emph{Information}, vol.~11, no.~2, p. 125, 2020.

\bibitem{he2016deep}
K.~He, X.~Zhang, S.~Ren, and J.~Sun, ``Deep residual learning for image recognition,'' in \emph{Proceedings of the IEEE Conference on Computer Vision and Pattern Recognition (CVPR)}, 2016, pp. 770--778.

\bibitem{deng2009imagenet}
J.~Deng, W.~Dong, R.~Socher, L.-J. Li, K.~Li, and L.~Fei-Fei, ``{ImageNet}: A large-scale hierarchical image database,'' in \emph{Proceedings of the IEEE Conference on Computer Vision and Pattern Recognition (CVPR)}, 2009, pp. 248--255.

\bibitem{ioffe2015batchnorm}
S.~Ioffe and C.~Szegedy, ``Batch normalization: Accelerating deep network training by reducing internal covariate shift,'' in \emph{Proceedings of the 32nd International Conference on Machine Learning (ICML)}, 2015, pp. 448--456.

\bibitem{bodla2017softnms}
N.~Bodla, B.~Singh, R.~Chellappa, and L.~S. Davis, ``Soft-{NMS}: Improving object detection with one line of code,'' in \emph{Proceedings of the IEEE International Conference on Computer Vision (ICCV)}, 2017, pp. 5562--5570.

\bibitem{solovyev2021weighted}
R.~Solovyev, W.~Wang, and T.~Gabruseva, ``Weighted boxes fusion: Ensembling boxes from different object detection models,'' \emph{Image and Vision Computing}, vol. 107, p. 104117, 2021.

\bibitem{loshchilov2019adamw}
\BIBentryALTinterwordspacing
I.~Loshchilov and F.~Hutter, ``Decoupled weight decay regularization,'' in \emph{International Conference on Learning Representations (ICLR)}, 2019. [Online]. Available: \url{https://openreview.net/forum?id=Bkg6RiCqY7}
\BIBentrySTDinterwordspacing

\bibitem{loshchilov2017sgdr}
------, ``{SGDR}: Stochastic gradient descent with warm restarts,'' in \emph{International Conference on Learning Representations (ICLR)}, 2017.

\end{thebibliography}

\end{document}